\documentclass[letterpaper, 10 pt, conference]{ieeeconf}  

\IEEEoverridecommandlockouts                             

\usepackage{graphics} 
\usepackage{epsfig} 

\usepackage{times} 
\usepackage{amsmath} 
\usepackage{amssymb}  
\usepackage{booktabs,ragged2e}
\usepackage{multirow}
\usepackage[flushleft]{threeparttable}
\usepackage{xcolor}
\usepackage[english]{babel}
\usepackage{lipsum}
\usepackage{hyperref}

\usepackage{hhline}
\usepackage{array}
\usepackage{cite}
\usepackage{filecontents}
\usepackage{url}
\usepackage{tabularx}

\usepackage{color}
\newcommand{\col}{\color{black}}
\newcommand{\colb}{\color{black}}
\newcommand{\colbb}{\color{black}}

\usepackage[ruled,linesnumbered]{algorithm2e}
\SetKwInput{KwRequire}{Require}

\title{\LARGE \bf
	Enhancing Exploration Efficiency \\using  Uncertainty-Aware Information Prediction
}

\author{Seunghwan Kim$^{1*}$, Heejung Shin$^{1*}$, Gaeun Yim$^{1}$, Changseung Kim$^{1}$ and Hyondong Oh$^{1}$
	\thanks{}
	\thanks{$^{1}$Department of Mechanical Engineering, Ulsan National Institute of Science and Technology (UNIST), Ulsan 44919, Republic of Korea
		{\tt\small kevin6960@unist.ac.kr, godhj@unist.ac.kr, gaeungraceyim@unist.ac.kr,  pon02124@unist.ac.kr, h.oh@unist.ac.kr}}%
        \thanks{*The first two authors contributed equally to this work.}%
}

\begin{document}
    \maketitle
	\thispagestyle{empty}
	\pagestyle{empty}	
	\begin{abstract}
        Autonomous exploration is a crucial aspect of robotics, enabling robots to explore unknown environments and generate maps without prior knowledge. This paper proposes a method to enhance exploration efficiency by integrating neural network-based occupancy grid map prediction with {\colbb uncertainty-aware} Bayesian neural network. Uncertainty from neural network-based occupancy grid map prediction is probabilistically integrated into {\colb mutual information for exploration}. To demonstrate the effectiveness of the proposed method, we conducted comparative simulations within a frontier exploration framework in a realistic simulator environment against various information metrics. The proposed method showed superior performance in terms of exploration efficiency.

        \end{abstract}
	
\section{Introduction}

    Autonomous exploration is a key research topic in robotics, aiming to enable robots to autonomously explore and navigate unknown environments without prior knowledge. For effective robot operation, accurate mapping of the environment is essential, as it allows the robot to move safely and efficiently. Exploration plays a critical role in enabling robots to autonomously generate these maps, avoid obstacles, and locate targets.

    For efficient exploration, it is necessary to determine and select the best sensing points that can provide the maximum coverage of the environment. To determine the best sensing points, it is essential to accurately evaluate the expected coverage at the next sensing point, which in autonomous exploration is generally referred to information. In the field of exploration, typical methods for evaluating the information include volumetric gain and mutual information. Volumetric gain is defined as the area of unknown regions that can be observed from the next sensing point\cite{bircher2016receding}. The volumetric gain is calculated by counting the number of unknown grids that a ray passes through after raycasting from a sensing point. This method is simple and quick to compute, making it the most widely used approach in the field of autonomous exploration. In contrast, in the mutual information, the information is probabilistically evaluated, taking into account the uncertainty of both the currently estimated map and the sensors. The mutual information is defined as the expected reduction in entropy of a map when a new sensor measurement is made from a sensing point. Calculating mutual information requires the computation of expected values, which can be time-consuming. Despite this, it offers the advantage of being able to consider the uncertainty of both the currently estimated map and the sensor measurements when evaluating information\cite{julian2014mutual}. To address the computational burden of calculating the mutual information, Henderson et al.\cite{charrow2015information} introduced Cauchy-Schwarz quadratic mutual information and Zhang et al.\cite{zhang2020fsmi} proposed the uniform fast Shannon mutual information (FSMI).

    Both volumetric gain and mutual information require predicting future measurements to evaluate the information. {\colb However, the problem arises when meaningful prediction of future measurements in unknown areas are unattainable. Unknown grids, each with a 0.5 probability of being occupied, signify an absence of information regarding whether the grid is occupied or free. Consequently, predicted future measurements for unknown areas can greatly differ from reality, leading to inaccurate assessments of the information.} Ultimately, to accurately assess information from the next sensing point, it is necessary to predict the unknown areas near the next sensing point. With this background, recent years have seen research efforts utilizing neural network-based occupancy grid map (OGM) prediction for exploration. Shrestha et al.\cite{shrestha2019learned} used a variational autoencoder-based occupancy map prediction to assess information at frontiers, demonstrating improvements in exploration performance. Ramakrishnan et al.\cite{ramakrishnan2020occupancy} proposed an autoencoder-based OGM prediction using RGB and depth images, and validated its effectiveness within a reinforcement learning-based exploration framework. In the research by Tao et al.\cite{tao2023seer}, occupancy prediction was extended to three-dimensional exploration using drones. Although neural network-based OGM prediction has been shown to enhance the efficiency of exploration, the extent to which its prediction can be trusted remains a concern. Incorrect prediction can lead to inappropriate information assessment, which may in turn impair exploration performance. On the other hand, if we can assess the uncertainty of neural network prediction, we can utilize its prediction more reliably and reduce the risks associated with prediction errors. 
    
    \textcolor{black}{Uncertainty in machine learning can be largely categorized into two types: aleatoric and epistemic~\cite{kendall2017uncertainties}. Aleatoric uncertainty, also known as data uncertainty, arises from inherent noise in the data and cannot be reduced by collecting more data. In contrast, epistemic uncertainty, or model uncertainty, originates from a lack of knowledge about the model and can be reduced by gathering more data. To quantify model uncertainty in deep neural networks, Bayesian neural networks(BNNs) are commonly used, as the Bayesian framework allows for uncertainty estimation through a posterior distribution. The predictive distribution of a BNN can be formulated as $p(y|x,D) = \int p(y|x, w) p(w|D) dw$.
where \(y\) is the predicted output, \(x\) is the input, \(D\) is the observed data, and \(w\) represents the model parameters. However, a naive BNN is computationally expensive, making it unsuitable for use in embedded systems, such as those on mobile robots. To address this limitation, Gal et al.~\cite{gal2016dropout} propose Monte Carlo (MC) dropout as a practical approximation of BNNs. {\colb Neural networks with MC dropout layers practically enable} the ability to estimate the uncertainty while being much faster and more feasible for real-time applications.}

    {\colb In utilizing OGM prediction for exploration, it is crucial for the information to reflect both the uncertainty of the prediction and the expected coverage. Uncertainty-driven planner for exploration and navigation (UPEN) \cite{georgakis2022uncertainty} has made efforts to use model ensemble-based uncertainty evaluation and the variance of neural network prediction as the information. However, the information metric in the UPEN considers only the uncertainty from neural network prediction, without reflecting the expected coverage. In contrast, the study by Shrestha et al. \cite{shrestha2019learned}, which employs deterministic OGM prediction, focuses solely on evaluating expected coverage. Therefore, to account for both the uncertainty of prediction and expected coverage, we have introduced a combination of OGM prediction based on BNNs and the mutual information.}  In our approach, the OGM prediction is represented probabilistically using BNN, and this is used to predict the probability distribution of future measurements. Finally, information prediction that incorporates neural network uncertainty is performed in the form of mutual information. To demonstrate the effectiveness of the proposed method, we compared our approach with six information evaluation metrics, including the variance of neural network prediction, within the frontier exploration framework.

\section{Preliminaries} \label{sec2}

\subsection{Occupancy Grid Mapping}{
An occupancy grid map models the environment in a discretized form, where each grid indicates whether a specific area is occupied. Each grid is typically assumed to be probabilistically independent and is updated using a Bayesian filter as new sensor measurements are received. The map is defined by the set of random variables $M=\{M_1,…,M_K\}$, where $K$ is the total number of grids, and  $M_i\in\{0,1\}$ is a binary variable representing whether the $i$-th grid is occupied. Depth measurements are represented by the random variable $Z$, and the measurement obtained at time $t$ is denoted as $z_t=(z^1_t,...,z^{n_z}_t)$, where $n_z$ is the number of beams in the scan. The robot’s state at time $t$ is represented by  $x_{t}$, with $x_{1:t}$ and $z_{1:t}$ denoting the sequences of the robot's states and measurements up to time $t$. Given the independence of each grid, the posterior probability distribution of the map $P(M|z_{1:t},x_{1:t})$ can be expressed as:

\begin{equation}
P(M|z_{1:t},x_{1:t}) = \prod_{1\le i \le K}P(M_i|z_{1:t},x_{1:t}). \label{eq01}
\end{equation}
The update of the posterior probability distribution for each grid in the OGM can be performed using:
\begin{multline}
\frac{P(M_i=1|z_{1:t},x_{1:t})}{P(M_i=0|z_{1:t},x_{1:t})} \\ = \frac{P(M_i=1|z_{1:t-1},x_{1:t-1})}{P(M_i=0|z_{1:t-1},x_{1:t-1})}\cdot \frac{P(M_i=1|z_{t},x_{t})}{P(M_i=0|z_{t},x_{t})}. \label{eq02}
\end{multline}
In our paper, we will denote the posterior probability that the $i$-th grid is occupied as $o_i=P(M_i=1|z_{1:t},x_{1:t})$.

}

\subsection{Frontier Exploration}{
Frontier exploration \cite{yamauchi1997frontier} is one of the most widely used methods in the field of autonomous exploration. It enables autonomous mapping by repeatedly moving a robot to the frontier, which is defined as the boundary between free space and unknown space. In frontier exploration, if no more frontiers are detected, the exploration of the environment is considered complete, and the algorithm terminates. In this paper, frontier exploration is adopted as a framework to demonstrate the effectiveness of the proposed information prediction.

\begin{figure}[!t]
\begin{center}
\includegraphics[width=0.4\textwidth]{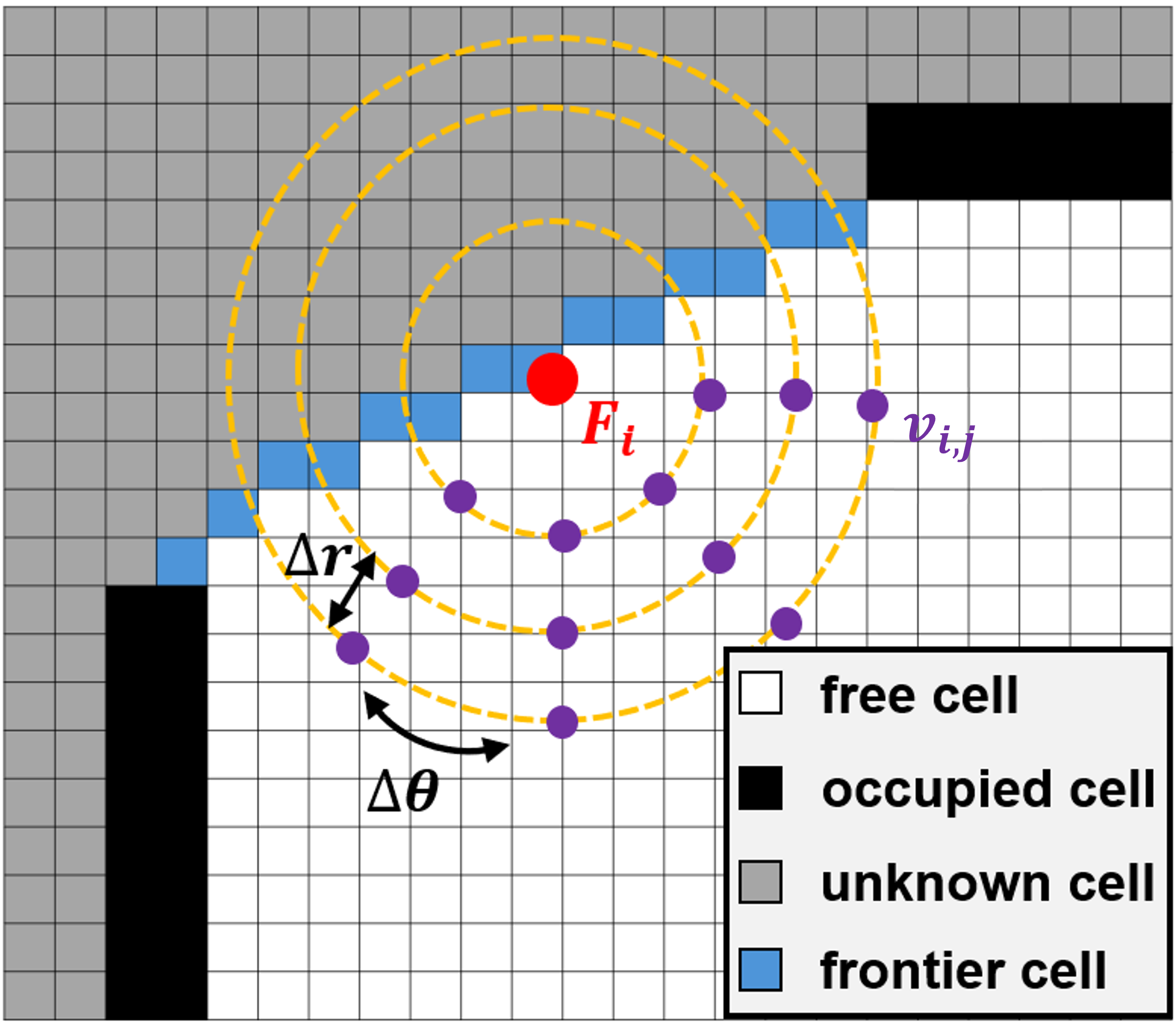}

\caption{Sample of the viewpoint generation. }

\label{fig1}
\end{center}
\end{figure}
    \subsubsection{Frontier Detection}
     Frontiers can be detected in an OGM by examining unknown grids adjacent to free grids. In this paper, the breadth-first search method is used to detect frontier grids\cite{keidar2012robot}. The detected frontier grids are then divided into separate frontiers based on continuity. Each frontier is represented by frontier centroid $F_i$, which is the central coordinate of the frontier grids within the frontier.    

    \subsubsection{Viewpoint Generation}
    After a frontier is detected, viewpoints that can observe the frontier are sampled. These viewpoints are uniformly sampled in a cylindrical coordinate system centered on the $F_i$, are considered only if they are located in free space. Assuming the use of a sensor with the proposed field of view, the $j$-th viewpoint for the $i$-th frontier $v_{i,j}$ is defined by a position and a yaw angle. The yaw angle is determined by selecting the direction that maximizes the information from that position, among eight directions spaced at equal intervals. And among these viewpoints, the one evaluated to have the highest information content is selected as the representative viewpoint $V_i$ for the frontier. An example of viewpoint generation is illustrated in Fig.~\ref{fig1}.
 
    \subsubsection{Select the Best Frontier}
    In frontier exploration, the frontier evaluated to have the highest utility is selected for navigation. By repeating this process, autonomous mapping is achieved. The utility considers the information at the frontier and the cost of distance to the frontier, and can be expressed as:
   
    \begin{equation}
    U(F_i)= I(V_i)\cdot \exp\left(-\lambda \cdot cost\left(x_t, V_i\right)\right), \label{eq_U}
    \end{equation}
    where $cost\left(x_t, V_i\right)$ is the path length between the current robot position $x_t$ and the viewpoint $V_i$, and $\lambda$ is the weight for the distance cost. The path length calculated using A* algorithm in the OGM. In this work, a $\lambda$ value of 0.05 was used. If the information $I(V_i)$ in the utility function is defined as $I_n(V_i)=1$, then frontier exploration becomes a nearest frontier exploration, where the robot moves to the nearest frontier. 
    
}

\subsection{Volumetric Gain}{
Volumetric gain from viewpoint $V$ can be defined as:

\begin{equation}
I_v(V) =\sum_{z^i \in z}^{} \sum_{M_j \in M^*}^{} I_u(M_j), \label{eq12}
\end{equation}

where 

\begin{equation}
I_u(M_j)=
\begin{cases}
1 & \text{if }  M_j = 0.5 , \\
0 & \text{otherwise} .
\end{cases}
\end{equation}
$z=\{z^1, z^2, ..., z^{nz}\}$ is a scan obtained by raycasting from viewpoint $V$ into the current OGM, where during raycasting, free and unknown grids are treated as passable by the ray, while occupied grids are treated as impassable. $M^{*}$ represents the set of grids in the OGM that the beam $z^i$ passes through. 

}

\subsection{Uniform FSMI}{
Let us consider $Z$ as the random variable representing depth measurements to be obtained in the viewpoint $V$. Assuming that each beam in the depth measurement is independent, the mutual information $MI$ can be expressed as:

\begin{equation}
I_m(V)=MI(M;Z)=\sum_{i=1}^{n_z} MI(M;Z^i), \label{eq04}
\end{equation}
where  $Z^i$ is a random variable representing $i$-th beam  of the measurement.
If we define the set of grids traversed by $Z^i$ as $M'=\{M_1,...,M_n\}$, then the mutual information for a single beam in \eqref{eq04} can be represented as:
\begin{multline}
 MI(M;Z^i)=MI(M';Z^i) =\sum_{j=1}^{n} MI(M_j;Z^i) \\ =\sum_{j=1}^{n} \int_z P(Z^i=z)f(\delta_j(z),r_j) dz, \label{eq05}
\end{multline}
where 
\begin{equation}
f(\delta,r)=\log \left(r+1/(r+\delta^{-1})\right)-\log\delta/(r\delta+1). 
\end{equation}
  $r_j=o_j/(1-o_j)$ is the odds ratio for the posterior probability of grid, and $\delta_j(z)$ represents the odds ratio of the inverse sensor model $\delta_j(z)$, expressed as:
\begin{multline}
\delta_j(z) = \frac{P(M_j=1|z_{t},x_{t})}{P(M_j=0|z_{t},x_{t})} \\ =
\begin{cases} 
\delta_{\text{occ}} & \text{if } z \text{ indicates } M_j \text{ is occupied}, \\
\delta_{\text{free}} & \text{if } z \text{ indicates } M_j \text{ is free}, \\
1 & \text{otherwise},\label{eq03}
\end{cases}
\end{multline}
where $\delta_{\text{occ}} > 1$ and $\delta_{\text{free}} < 1$ are are hyper parameters. $P(Z^i)$ in \eqref{eq05} can be represented as:
\begin{equation}
P(Z^i)=\sum_{k=0}^n P(e_k)P(Z^i|e_k),  \label{eq07}
\end{equation}
where $P(e_k)=o_k\prod_{i<k}(1-o_i)$ represents the probability that the $k$-th grid of $M'$ is initially occupied, while $P(e_0)$ represents the probability that all grids are free. Applying this, the mutual information for a single beam as in \eqref{eq05} can be expressed as:
\begin{equation}
I(M;Z^i)=\sum_{k=0}^n P(e_k) \int_z P(z|e_k) \left(\sum_{j=1}^{n} f(\delta_j(z),r_j)\right) dz.  \label{eq08}
\end{equation}
Given that $\sum_{j=1}^{n} f(\delta_j(z),r_j)$ is a piecewise constant summation that changes values only at the boundaries of each grid, we divide the integration over $z$ into a sum of multiple integration intervals across each grid boundary. Therefore, \eqref{eq08} can be expressed as:
\begin{multline}
MI(M;Z^i)
=\sum_{k=0}^n P(e_k) \sum_{m=1}^n  \int_{l_m}^{l_{m+1}} P(z|e_k) 
\\ \left(f(\delta_{occ},r_m)+\sum_{s<m} f(\delta_{emp},r_s)\right) dz \\
=\sum_{k=0}^n P(e_k) \sum_{m=1}^n \int_{l_m}^{l_{m+1}} P(z|e_k) C_m dz 
\\ =\sum_{k=0}^n \sum_{m=1}^n P(e_k) C_m \int_{l_m}^{l_{m+1}} P(z|e_k) dz,\label{eq09}
\end{multline}
where $l_m$ represents the distance from the sensing origin to the 
$m$-th grid of $M'$ and {\colb $C_m = f(\delta_{occ},r_m)+\sum_{s<m} f(\delta_{emp},r_s)$}.
In the Uniform FSMI, the sensor noise model is assumed to be a uniform distribution and can be expressed as:
\begin{equation}
P(Z^i|e_k)\sim U[l_k-H\Delta L,l_{k+1}-H\Delta L], \label{eq10}
\end{equation}
where $\Delta L$ is $l_{k+1}-l_k$ and $H \in \mathbb{Z}^{+}$ is scaling factor for the width of the uniform distribution.
When applying the sensor model from \eqref{eq10} to \eqref{eq09}, the final expression for the mutual information of a single beam can be represented as:
\begin{multline}
MI(M;Z^i)= \sum_{k=0}^n P(e_k)\sum_{m=k-H}^{k+H} \frac{C_m}{2H+1} \\ = \sum_{k=0}^n P(e_k)  \frac{D_{k+H}-D_{k-H-1}}{2H+1},\label{eq11}
\end{multline}
where $D_j=\sum_{m \le j} C_m$. More details about Uniform FSMI are described in \cite{zhang2020fsmi}.
}

\section{Information Prediction} \label{sec3}

\subsection{\textcolor{black}{Occupancy Grid Map Prediction Using Bayesian Neural Network}}{

\begin{figure*}[!ht]
    \centering
    \includegraphics[width=0.85\textwidth]{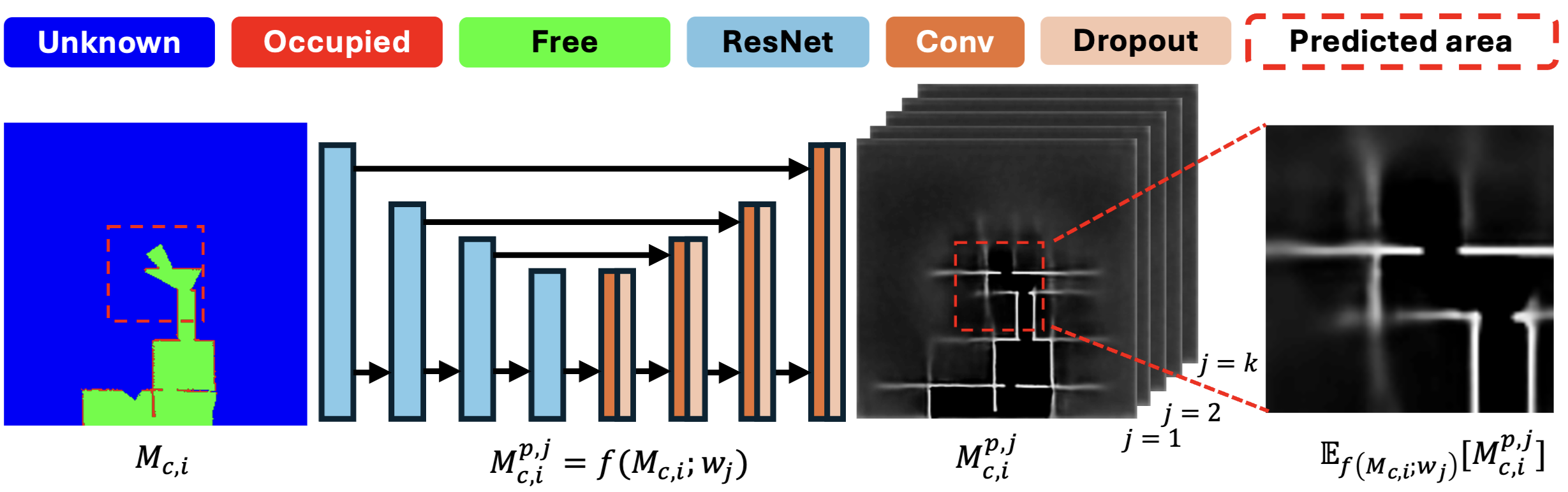}
    \caption{Occupancy grid map prediction {\colb at the $i$-th frontier} using a deep neural network. }
    \label{fig2}
\end{figure*}

{\col In our work, OGM prediction is defined as a binary semantic segmentation task, classifying unknown grids into occupied or free classes. To generate training data, we utilized the Housexpo dataset \cite{"li2019houseexpo"}, a 2D indoor layout dataset, to create 2,000 environments in GAZEBO. A turtlebot equipped with a depth camera explored environments using the nearest frontier algorithm to collect the data. During the data collection process for training, we crop the OGM and the ground truth map to a size of $256 \times 256$ pixels centered on each frontier centroid, with a resolution of 0.1m, to create the input data and target data, respectively. The input data $M_{c}\in \mathbb{R}^{4\times 256 \times 256}$ has 4 channels; free, occupied, unknown, and predicted area, as proposed by Shrestha et al.~\cite{shrestha2019learned} and the ground truth $M_{c}^{gt} \in \mathbb{R}^{1\times 256 \times 256}$ has a single channel representing occupancy probability. The dataset consists of 17,653 samples for training, 1,765 samples for validation, and 1,000 samples for testing.
}

We designed the U-Net based auto-encoder architecture of neural networks, with the encoder based on ResNet-18 and the decoder consisting of several convolution layers with MC dropout layers attached to each convolution layer allowing us to consider the model uncertainty. The framework for our deep neural network is depicted in Fig.~\ref{fig2}.
With this dataset $D$, we train the neural network using a batch size of 128, 100 epoch, a learning rate of 1e-3, a weight decay of 3e-5, and a dropout ratio of 0.2, along with simple data augmentation using random horizontal and vertical flips. Since we formulate the problem as a binary image segmentation task to predict the occupancy probability of each grid, we use binary cross-entropy loss. It is important to note that only the predictive area ($80\times80$ pixels) is considered for the loss calculation, and only the predicted area is utilized for robot exploration, as shown in Fig. ~\ref{fig2}.}

\textcolor{black}{During inference, when the input data $M_{c,i}$ is given {\colb for $i$-th frontier}, the prediction of the neural network is performed $n_s$ times. The probability of the prediction map at the i-th frontier $M^p_{c,i}$ can be approximately computed as:}
\begin{equation}
P(M^p_{c,i}|M_{c,i}, D) \approx \frac{1}{n_s}\sum_{j=1}^{n_s} f(M_{c,i};w_j), \label{eqnn}
\end{equation}
\textcolor{black}{where $w_j$ represents the sampled weights of the BNN, and in this work, $n_s$ was set to 10. Additionally, the prediction map $P(M^p)$ is generated by replacing the unknown grids in the area corresponding to $M_{c,i}$ in $P(M|z_{1:t},x_{1:t})$ with the corresponding values from $P(M^p_{c,i}|M_{c,i}, D)$ {\colb for all frontiers}.}
The variance of neural network prediction can be expressed as:
\begin{equation}
Var(M^p_{c,i}) \approx \frac{1}{n_s}\sum_{j=1}^{n_s} f(M_{c,i};w_j)^2 -\left(\frac{1}{n_s}\sum_{j=1}^{n_s} f(M_{c,i};w_j)\right)^2. \label{eqnn2}
\end{equation}
The variance map $Var(M^p)$ contains values for $Var(M^p_{c,i})$ corresponding to $M_{c,i}$ in 
$K$ grids, and grids where OGM prediction has not been performed have a value of zero.

\subsection{Information Metric with Map Prediction}{

 This section introduces four information metrics utilizing OGM prediction. {\colbb The volumetric gain with map prediction evaluates information by deterministically using OGM prediction, following the approach used by Tao et al.~\cite{tao2023seer}. Two additional metrics, inspired by the UPEN approach, are presented, which utilize the variance of OGM prediction, each differing in their methods of sensor visibility evaluation. Lastly, our proposed method, the uniform FSMI with map prediction, integrates the uncertainty of neural network prediction into the information assessment by using a probabilistically evaluated prediction map.} 

\subsubsection{Volumetric Gain with Map Prediction}
Volumetric gain does not use probabilistic concepts for information evaluation, which means that map prediction is used deterministically regardless of the uncertainty in neural network inference. The occupied or free grid of the prediction map is classified based on a probability threshold of 0.5; Values greater than 0.5 in the prediction map are considered occupied, otherwise they are considered free. The volumetric gain with map prediction is expressed as:
\begin{equation}
PI_v(V) =\sum_{z_p^i \in z_p}^{} \sum_{M_j \in M^{*}}^{} I_u(M_j),
\end{equation}
where $z_p$ is a scan obtained by raycasting from viewpoint $V$ into the prediction map, where during raycasting and $M^*$ represents the set of grids in the current OGM that the beam $z_p^i$ passes through.

\subsubsection{Variance of Prediction Map}
{\colbb In UPEN~\cite{kendall2017uncertainties}, the variance of neural network inference was used as an information metric, but sensor visibility was not considered. While the assumption that the variance of prediction map correlates strongly with information is reasonable, information should be evaluated based on areas visible from the next sensing point. Therefore, we use the variance of prediction map as an information metric, but only within areas where sensor visibility is guaranteed from viewpoint $V$, which can be expressed as:}
\begin{equation}
PI_{var1}(V) =\sum_{z^i \in z}^{} \sum_{M^p_j \in M^p{*}}^{} Var(M^p_j), \label{eqv1}
\end{equation}
where {\colbb$z^i$ is obtained by performing raycasting on the current OGM, as in \eqref{eq12},} and $M^{p*}$ represents the set of grids in the prediction map that the beam $z^i$ passes through. 

\subsubsection{Variance of Prediction Map with Predicted Raycasting}
{\colb In  \eqref{eqv1}, the visibility was evaluated based on the current OGM, but the prediction map can also be utilized for evaluating visibility. When considering the prediction map for visibility, it can be expressed as:}
\begin{equation}
PI_{var2}(V) =\sum_{z_p^i \in z_p}^{} \sum_{M^p_j \in M^p{*}}^{} Var(M^p_j).\label{eqv2}
\end{equation}
}{\colb Note that $z_p$ represents a scan obtained by raycasting on the prediction map.}

\subsubsection{Uniform FSMI with Map Prediction}
\begin{figure}[!t]
\begin{center}
\includegraphics[width=0.475\textwidth]{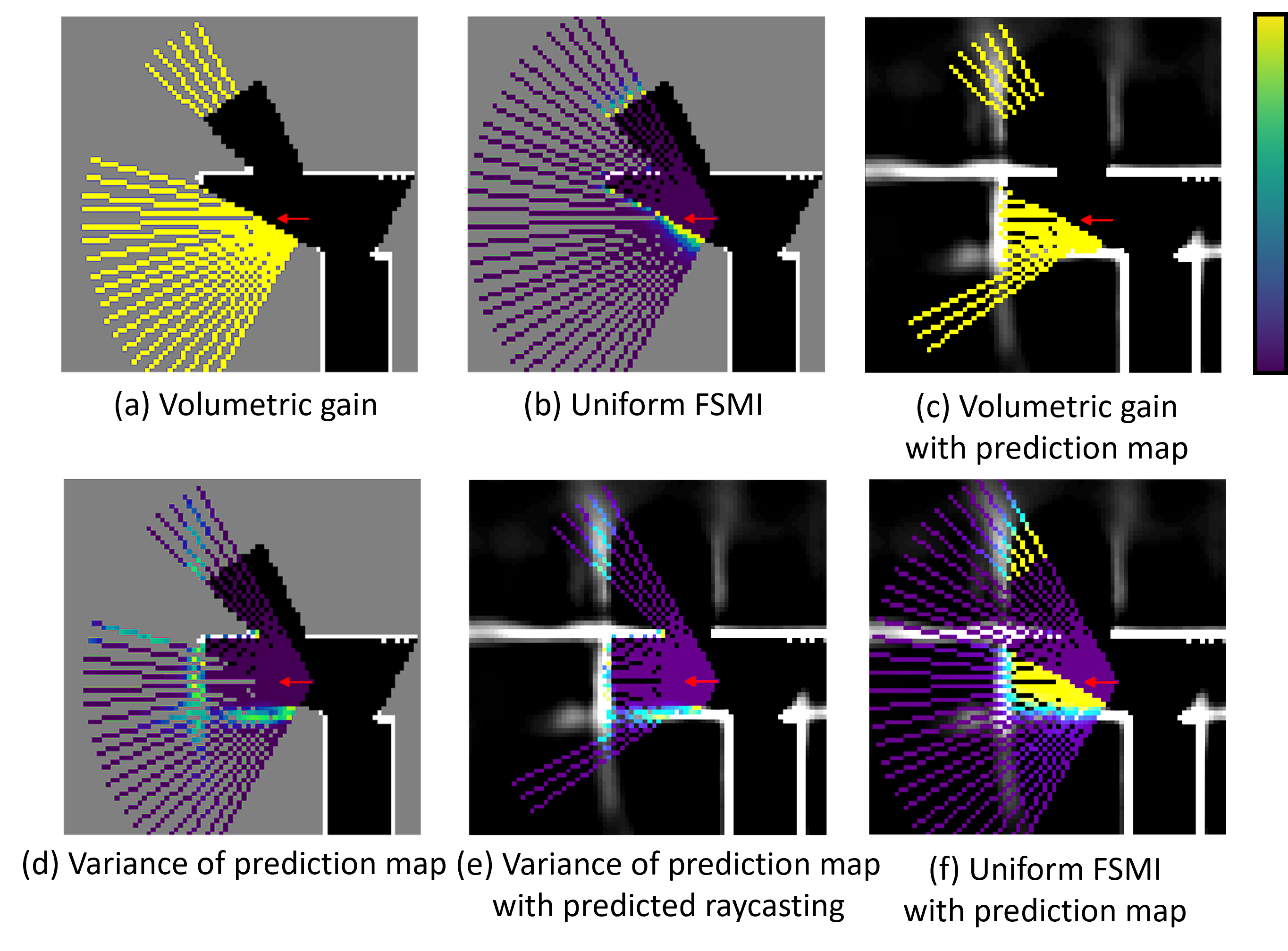}
\caption{Illustrative results for each information metric. {\colb The red arrow indicates the position and direction of the sensor. In the uniform FSMI and variance of prediction map, colors closer to yellow represent higher information, while colors closer to purple indicate lower information.}}
\label{fig3}
\end{center}
\end{figure}

In mutual information, future measurement prediction is probabilistically modeled, allowing the uncertainty of neural network inference to be incorporated into information evaluation based on this. Based on the prediction map, the probability of the $i$-th beam in future measurements can be expressed as:
\begin{equation}
P(Z_p^i)=\sum_{k=0}^n P(e^p_k)P(Z_p^i|e^p_k),
\end{equation}
where $P(e^p_k)=o^p_k\prod_{i<k}(1-o^p_i)$ and $o^p_i=P(M^p_i=1)$.
Using this, the uniform FSMI utilizing map prediction can be expressed as: 
\begin{equation}
PI_m(V)=\sum_{i=1}^{n_z} MI(M;Z_p^i),
\end{equation}
where
\begin{equation}
 MI(M;Z_p^i)= \sum_{k=0}^n P(e^p_k)  \frac{D_{k+H}-D_{k-H-1}}{2H+1}.
\end{equation}
Figure~\ref{fig3} represents example results for each information metric.

\section{Experimental Validation}

\subsection{Occupancy Grid Map Prediction Results}

We validate the performance of using the proposed neural network's OGM prediction on the test dataset. The test set consists of 1,000 input and ground truth pairs, which are independent of the training and validation datasets. Table~\ref{ogm_prediction} summarizes the performance of the network. We adopt the metrics commonly used for segmentation tasks. Latency is evaluated on an NVIDIA RTX 3090 GPU using 10 iterations to approximate the posterior in \eqref{eqnn}.
\begin{table}[!t]
\caption{OGM prediction performance}
\begin{center}
\begin{tabular}{c c c c c c}
\hline
 \textbf{\textit{Accuracy}} & \textbf{\textit{Precision}} & \textbf{\textit{Recall}} & \textbf{\textit{F1}} & \textbf{\textit{IoU}}  & \textbf{\textit{Latency (ms)}}   \\
\hline
 0.980 & 0.815 & 0.554 & 0.636 & 0.518 & 25.7 \\
\hline
\end{tabular}
\label{ogm_prediction}
\end{center}
\end{table}

\subsection{Autonomous Exploration Simulation with Occupancy Grid Map Prediction}{
The simulation was conducted using a mobile sensor setup consisting of a turtlebot and a depth camera in a GAZEBO environment. The depth camera was assumed to have a maximum range of 5 meters and a 90-degree field of view. The simulation was carried out in four different environments as shown in Fig.~\ref{fig4}. Based on the depth camera data, OGM was generated according to (2), and the robot moved to the frontier with the highest utility. Methods utilizing seven different pieces of information—$I_v$, $I_m$, $I_n$, $PI_v$, $PI_m$, $PI_{var1}$, and $PI_{var2}$—were applied to calculate the utility function in (4). Additionally, the results of these methods were represented in Table~\ref{tab0} and Fig.~\ref{fig5}. The values in Table~\ref{tab0} represent the average exploration time from 10 experiments, with the standard deviations indicated in parentheses. Exploration time refers to the duration taken to complete the exploration of an environment.

The results in Table~\ref{tab0} indicate that the combination of the prediction map and uniform FSMI consistently shows the shortest average exploration time across all environments. Notably, this approach also demonstrates the smallest standard deviation in exploration times among methods utilizing the prediction map, indicating a stable performance.

\begin{figure}[!h]
\begin{center}
\includegraphics[width=0.45\textwidth]{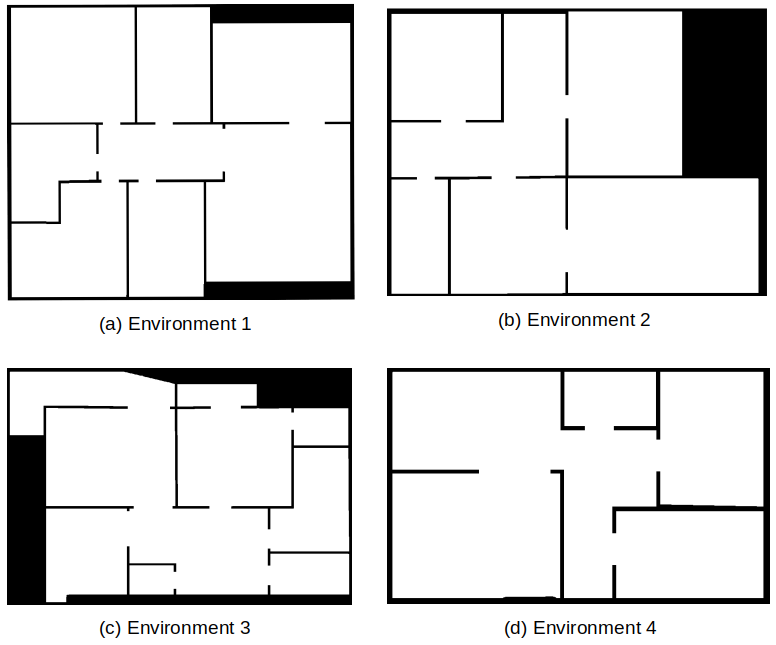}
\caption{Simulation environments.}
\label{fig4}
\end{center}
\end{figure}

\begin{figure}[!t]
\begin{center}
\includegraphics[width=0.475\textwidth]{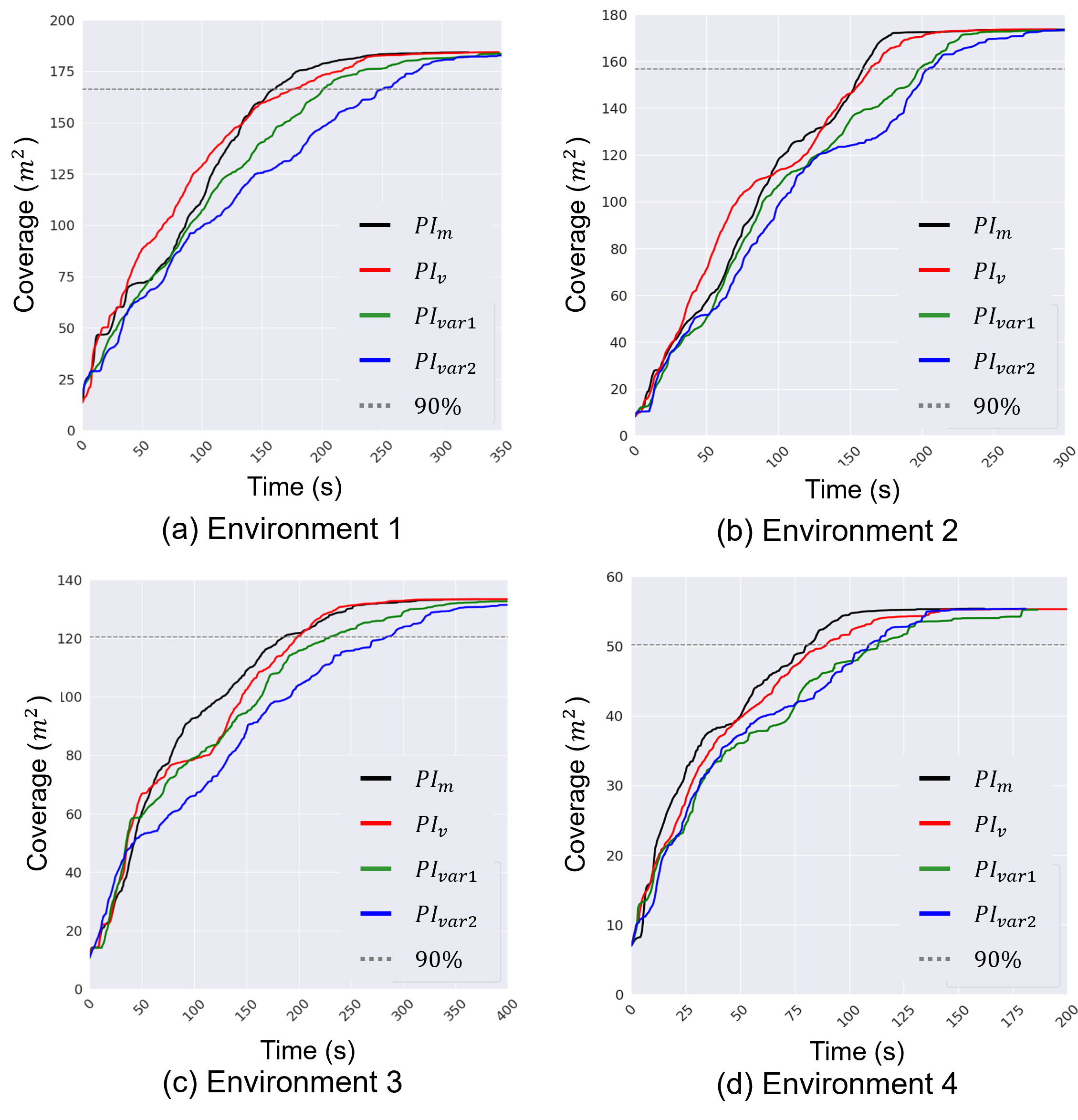}
\caption{Autonomous exploration simulation results. Each plot represents the average of 10 trials.}
\label{fig5}
\end{center}
\end{figure}
}

 \begin{table}[!t]
 \setlength{\tabcolsep}{2pt}
\caption{Result in the autonomous exploration simulation }

\begin{center}
\begin{tabular}{c| c c | c| c c}
\hline

\textbf{\textit{Env}}&\textbf{\textit{method}}& \textbf{\textit{Exploration Time [s]}}  &\textbf{\textit{Env}}&\textbf{\textit{method}}& \textbf{\textit{Exploration Time [s]}}\\
\hline
\hline
& $PI_m$ & \textbf{157.54} (18.30) & & $PI_m$ &  \textbf{155.68} (9.76) \\
& $PI_v$ & 174.42 (31.57) &&$PI_v$ & 160.82 (15.73) \\
 & $PI_{var1}$ & 201.12(25.80) &&$PI_{var1}$ & 190.85 (21.69)\\
1 & $PI_{var2}$ & 236.53(29.96) & 2&$PI_{var2}$ & 210.65 (26.19)\\
& $I_m$ & 269.91 (17.44) &&$I_m$ & 210.33 (21.16)\\
& $I_v$ & 233.11 (16.79) &&$I_v$ & 171.12 (10.81)\\
& $I_n$ & 282.33 (55.27) &&$I_n$ & 208.32 (26.19)\\
\hline
& $PI_m$ &  \textbf{183.09} (14.78) & & $PI_m$ &  \textbf{77.64} (9.02) \\
& $PI_v$ & 195.92 (18.71) &&$PI_v$ & 90.82 (19.74) \\
 & $PI_{var1}$ & 229.16 (40.52) &&$PI_{var1}$ & 113.21 (31.56)\\
3 & $PI_{var2}$ & 266.38 (32.80) & 4&$PI_{var2}$ & 111.49 (13.23)\\
& $I_m$ & 227.30 (23.93) &&$I_m$ & 98.18 (17.61)\\
& $I_v$ & 208.67 (27.26) &&$I_v$ & 97.18 (16.02)\\
& $I_n$ & 248.69 (10.82) &&$I_n$ & 109.88 (19.83)\\
\hline
\end{tabular}
\label{tab0}
\end{center}
\end{table}

\section{Conclusions and Future Work}
In this work, we conducted neural network-based prediction of OGM and quantified the uncertainty of its prediction in probabilistic form using BNN. In the evaluation of mutual information, the prediction of future measurements was based on the probabilities derived from the neural network OGM prediction, integrating the uncertainty of neural network prediction into the information evaluation process. To demonstrate the effectiveness of our proposed method, we compared it against the deterministic use of neural network prediction based on volumetric gain and methods that use neural network uncertainty defined by variance as information. Our comparisons were conducted within the frontier exploration framework through simulations in the GAZEBO environment. In four different environments, the proposed method demonstrated higher exploration efficiency compared with other methods.

However, our work has several limitations, and we plan to expand in the following directions. First, while we have validated the proposed method in a realistic simulator, validation in the real world remains a crucial aspect of autonomous exploration. Therefore, we will validate the proposed method through real-world experiments.
Despite approximating the posterior probability distribution of neural network prediction using MC dropout, the latency caused by multiple inferences poses a limitation for practical use. Therefore, we plan to develop models that can be practically used in embedded systems by utilizing neural network compression.
Additionally, the uncertainty in neural network prediction arises from two sources: epistemic uncertainty which stems from the model's lack of knowledge and aleatoric uncertainty, which originates from the inherent noise in the data itself. Our current uncertainty assessment, based on BNN, primarily addresses epistemic uncertainty. However, aleatoric uncertainty is also a significant factor in OGM prediction. Thus, we plan to extend our uncertainty evaluation to include aleatoric uncertainty as well.

	\addtolength{\textheight}{-12cm}   
	
	\bibliographystyle{IEEEtran}
	\bibliography{references.bib}
	
\end{document}